\documentclass[conference]{IEEEtran}
\usepackage{amsmath,epsfig}
\usepackage{bbold}
\usepackage{graphicx,amsfonts,amssymb,amsmath,nicefrac}
\usepackage{fixmetodonotes}
\usepackage{xcolor}

\usepackage[all=normal,bibliography=tight,mathspacing=tight,paragraphs=tight,bibbreaks=tight,wordspacing=tight]{savetrees}
\usepackage{enumitem}

\begin{document}

% Example definitions.
% --------------------
\def\x{{\mathbf x}}
\def\L{{\cal L}}

% Title.
% ------
\title{Self-Supervised Multimodal Domino: in Search of Biomarkers for Alzheimer's Disease}
%
% Single address.
% ---------------
\author{
  \IEEEauthorblockN{
    Alex Fedorov\IEEEauthorrefmark{1}\IEEEauthorrefmark{4},
    Tristan Sylvain\IEEEauthorrefmark{5}\IEEEauthorrefmark{6},
    Eloy Geenjaar\IEEEauthorrefmark{4}\IEEEauthorrefmark{7},
    Margaux Luck\IEEEauthorrefmark{5}, \\
    Lei Wu\IEEEauthorrefmark{2}\IEEEauthorrefmark{4},
    Thomas P. DeRamus\IEEEauthorrefmark{2}\IEEEauthorrefmark{4},
    Alex Kirilin,
    Dmitry Bleklov, \\
    Vince D. Calhoun\IEEEauthorrefmark{1}\IEEEauthorrefmark{2}\IEEEauthorrefmark{3}\IEEEauthorrefmark{4} and
    Sergey M. Plis\IEEEauthorrefmark{2}\IEEEauthorrefmark{4}
  }
  \IEEEauthorblockA{\IEEEauthorrefmark{1}Georgia Institute of Technology,  \IEEEauthorrefmark{2}Georgia State University, \IEEEauthorrefmark{3}Emory University}
  \IEEEauthorblockA{\IEEEauthorrefmark{4}Center for Translational Research in Neuroimaging and Data Science, Atlanta, GA, USA}
  \IEEEauthorblockA{
    \IEEEauthorrefmark{5}Mila - Quebec AI Institute, \IEEEauthorrefmark{6}Université de Montréal, Montreal, Quebec, Canada
  }
  \IEEEauthorblockA{
    \IEEEauthorrefmark{7}Delft University of Technology, Delft, the Netherlands
  }
}

\maketitle
\begin{abstract}
  Sensory input from multiple sources is crucial for robust and coherent human perception. Different sources contribute complementary explanatory factors. Similarly, research studies often collect multimodal imaging data, each of which can provide shared and unique information. This observation motivated the design of powerful multimodal self-supervised representation-learning algorithms. In this paper, we unify recent work on multimodal self-supervised learning under a single framework. Observing that most self-supervised methods optimize similarity metrics between a set of model components, we propose a taxonomy of all reasonable ways to organize this process. We first evaluate models on toy multimodal MNIST datasets and then apply them to a multimodal neuroimaging dataset with Alzheimer's disease patients. We find that (1) multimodal contrastive learning has significant benefits over its unimodal counterpart, (2) the specific composition of multiple contrastive objectives is critical to performance on a downstream task, (3) maximization of the similarity between representations has a regularizing effect on a neural network, which can sometimes lead to reduced downstream performance but still reveal multimodal relations. Results show that the proposed approach outperforms previous self-supervised encoder-decoder methods based on canonical correlation analysis (CCA) or the mixture-of-experts multimodal variational autoEncoder (MMVAE) on various datasets with a linear evaluation protocol. Importantly, we find a promising solution to uncover connections between modalities through a jointly shared subspace that can help advance work in our search for neuroimaging biomarkers.
\end{abstract}

\section{Introduction}
\label{sec:intro}

% multimodal data...
% {\color{red}TODO multimodal data}

The idealized tasks on which machine learning models are benchmarked commonly involve a single data source and readily available labels. Many datasets, however,
are composed of multiple sources, e.g., different MRI modalities~\cite{calhoun2016multimodal, fedorov2020self} in medical imaging, LiDAR, and video for self-driving cars~\cite{xiao2020multimodal}, and confounder-influenced data~\cite{scholkopf2016modeling}. Additionally, labels can be scarce, incorrect, or too definitive, which leads to the need for self-supervised or at least semi-supervised learning. In this work, we will be addressing both these constraints simultaneously by proposing an analysis of self-supervised learning approaches to multimodal data and comparing them to more classical methods. The self-supervised learning methods we consider are based on contrastive Deep InfoMax (DIM)~\cite{dim}.

Classical approaches to multimodal data include \emph{canonical correlation analysis} (CCA)~\cite{hotelling1992relations}, which finds maximally correlated linear projections of two data sources. More recently, CCA has been extended to allow for representations obtained using neural networks in works such as \emph{deep canonical correlation analysis} (DCCA)~\cite{andrew2013deep} and \emph{deep
canonically correlated autoencoders} (DCCAE)~\cite{wang2015deep}. Another family of methods is based on variational autoencoders, such as the \emph{multimodal mixture of experts VAE} (MMVAE)~\cite{shi2019variational}, which has resulted in considerable performance improvements.

\begin{figure}[t]
  \center
  \includegraphics[width=\linewidth]{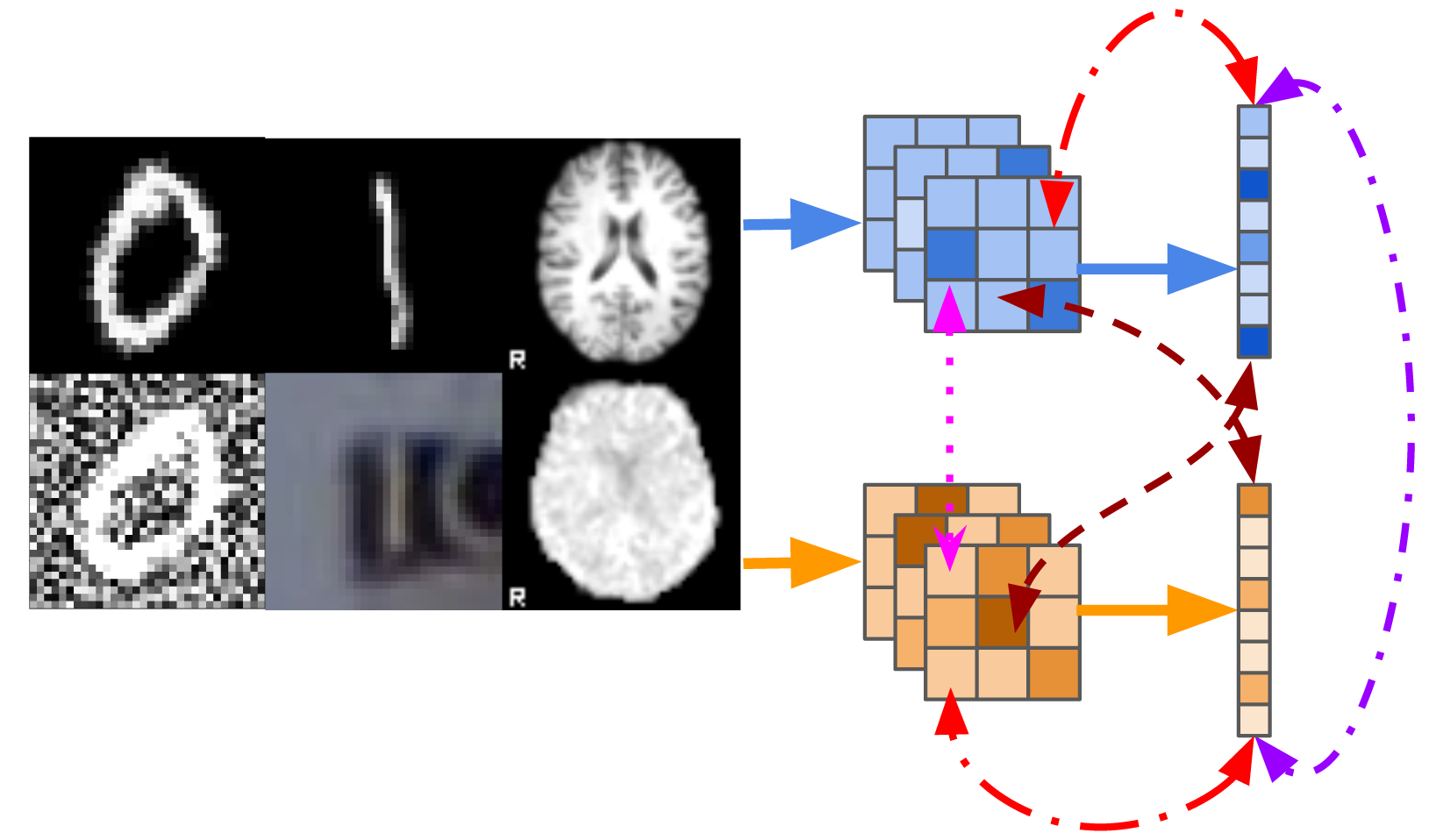}
  \caption{Sample images from Two-View MNIST, MNIST-SVHN, OASIS-3, and a general scheme of all the possible ways to maximize the mutual information between two modalities. The arrows represent all possible combinations as coordinated connections between vectors of locations (along convolutional channels) or whole latent representations. The red arrow is Convolution-to-Representation (CR), the pink --- Convolution-Convolution (CC), the burgundy --- Cross-Convolution-to-Representation (XX), and the purple --- Representation-to-Representation (RR).}
  \label{fig:dataset_scheme}
\end{figure}

Contrastive objectives have in recent years become essential components for a large number of self-supervised learning methods. Mutual information estimation~\cite{belghazi2018mine} has inspired a number of successful uses for single-view data, such as \emph{deep infomax} (DIM)~\cite{hjelm2018learning}, \emph{contrastive predictive coding} (CPC)~\cite{oord2018representation}) and for multi-view data, such as \emph{augmented multiscale DIM} (AMDIM)~\cite{bachman2019amdim}, \emph{contrastive multiview coding} (CMC)~\cite{tian2019contrastive}, SimCLR~\cite{chen2020simple}) image classification, reinforcement learning (\emph{spatio-temporal DIM} (ST-DIM)~\cite{anand2019unsupervised}) and zero-shot learning (\emph{class matching DIM} (CM-DIM)~\cite{sylvain2019locality,sylvain2020zeroshot}). Such methods have resulted in large representation learning improvements by considering different views of the same instance. These and other related methods mostly operate in a self-supervised fashion, where the goal is to encourage similarity between transformed representations of a single instance. These objectives can also be readily applied to a multimodal context, where different sources can be understood as different views of the same instance.
In addition to the relative scarcity of the literature on self-supervised multimodal representation learning, we also note that there are no studies that consider explicit combinations of unsupervised and multimodal objectives. This work aims to contribute to both issues. Further, we propose a taxonomy for self-supervised learning that is readily applicable to newer models: contrasting with momentum encoders MoCo~\cite{he2020momentum}, prototypical/clustering approaches SwAV~\cite{caron2020unsupervised}, PCL~\cite{li2020prototypical}, and non-contrastive approaches, such as BYOL~\cite{grill2020bootstrap} and SimSiam~\cite{chen2020exploring}.

Our contributions are as follows:
\begin{itemize}
    \item We unify contrastive learning methods under a systematic framework and perform a comparison of their merits. By analyzing the effect of adding different contrastive objectives, we show that correctly combining such objectives significantly impacts performance.
    \item We demonstrate empirically that multimodal contrastive learning has significant benefits over its unimodal counterpart.
    \item We propose novel models that emerge naturally from our systematic framework and taxonomy.
    \item We show that simultaneously maximizing the inter- and intra-modality mutual information is essential because intra-modality similarity maximization alone may lead to a collapse of the representations, which could, in turn, be caused by underspecification~\cite{d2020underspecification}.
    \item We propose a new analysis with \emph{centered kernel alignment} (CKA)~\cite{kornblith2019similarity} to test the alignment of representations between modalities as a measure of their joint shared subspace.
    \item We propose solutions to the possible detrimental effects of a similarity metric on representations.
    \item We applied modern contrastive techniques to a complex multimodal medical setting and found a promising solution to uncover connections between modalities through a jointly shared subspace.
\end{itemize}

% Canonical correlation analysis

% Not all well-performing contrastive objectives rely on mutual information: BYOL~\cite{grill2020bootstrap} achieves strong performance on image classification, and OC-GAN~\cite{sylvain2020objectcentric} uses a ranking-based objective for image generation.

% {\color{red}TODO BYOL~\cite{grill2020bootstrap}}

\begin{figure}[t]
  \center
  \includegraphics[width=1\linewidth]{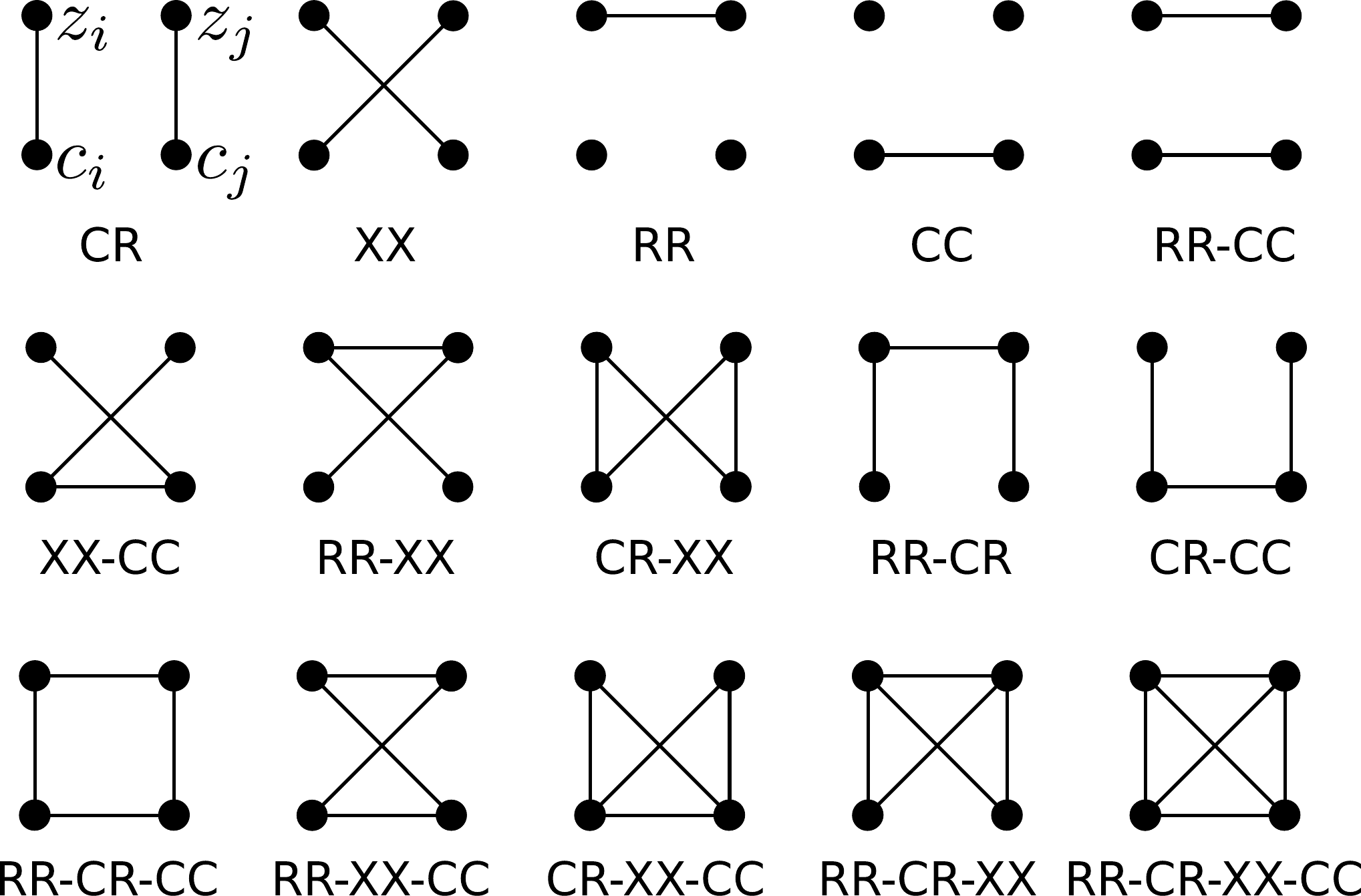}
  \caption{Possible variants of connected pairs to maximize mutual information. The 4 main objectives are CR, XX, CC, RR. CR (Convolution-to-Representation) is a unimodal objective capturing the relation between the location in the Convolutional feature map and the Representation. XX corresponds to a relation between a location in the convolutional feature map of one modality and the representation of the other. CC (Convolution-to-Convolution) relates locations in the Convolutional feature map across modalities. RR captures a multimodal Representation-to-Representation relation. The other objectives are combinations of the first four (CR, XX, RR, and CC).}
  \label{fig:variants}
\end{figure}

\begin{figure}[t]
  \center
  \includegraphics[width=\linewidth]{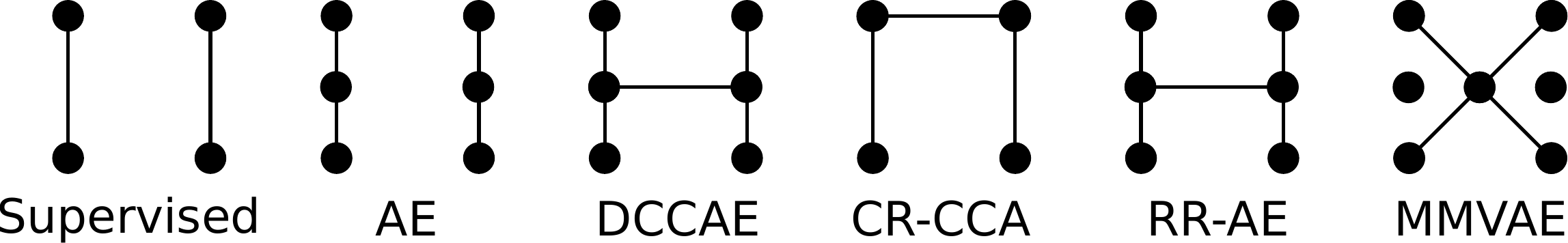}
  \caption{Scheme for Supervised, AE, DCCAE, L-CCA, RR-AE and MMVAE.}
  \label{fig:others}
\end{figure}

\section{Methods}

\subsection{Problem setting}

Let $\{D_i = \{x_1,...,x_N\}\}_{i=1,...,n}$ be a set of datasets $D_i$ with $N$ samples and $n$ modalities. For each $i$th modality $D_i$, we define a sampled image $x_i$, a CNN encoder $E_i$, a convolutional feature $c_i$ from a fixed layer $l$ in the encoder as $c_i = E^l_i(x_i)$, which is needed to define DIM~\cite{dim}-based objective, and a latent representation $z_i$ defined as $z_i = E_i(x_i)$. The AE-based approaches also produce a reconstruction $x'_i$ of the original sample $x_i$.

To learn the set of encoders $\{E_i\}_{i=1,...,n}$ we want to maximize the following objective $L$:
\begin{equation*}
  \mathcal{L} = \sum_{i=1}^n \sum_{j=1}^n \ell(D_i,D_j),
\end{equation*}
where $\ell(D_i,D_j)$ is a loss function between datasets $D_i$ and $D_j$.
In this study, we are specifically exploring the self-supervised contrastive objectives based on the maximization of mutual information, $\ell(D_i, D_j)$ is therefore a loss function that maximizes the mutual information between the datasets of the two modalities.

\subsection{Contrastive mutual information maximization}

To maximize the lower bound of the mutual information, we utilize the InfoNCE~\cite{oord2018representation} estimator $I$. We chose InfoNCE, because it is the common choice in the literature for self-supervised learning in natural images. There are, however, multiple estimators available (e.g., JSD and NWJ)~\cite{dim, tschannen2019mutual, poole2019variational}. We define $I$ by adopting implementation from AMDIM~\cite{bachman2019amdim}:
\begin{align*}
  &\ell(D_i,D_j) = I(D_i; D_j) \\
  &\ge \frac{1}{N} \sum_{l=1}^N \log \frac{e^{f(u^l_i,v^l_j)}}{\sum_{k=1, k\ne l}^N  e^{f(u^l_i,v^k_j)}  + e^{f^{c}(u^l_i,v^l_j)}},
\end{align*}
where $f$ is a critic function, $u_i, v_j$ are two embeddings. The embeddings $u_i,v_j$ are obtained by projecting the location of a convolutional feature $c$ or a latent $z$ (e.g. $u_i = \phi(c_i)$ and $v_j = \psi(z_j)$). The neural networks that parametrize these projections $\phi$ and $\psi$ are also known as projection heads~\cite{chen2020simple}.

To describe a critic function, we define positive and negative pairs. A pair $(u_i, v_j)$ is positive if it is sampled from a joint distribution $(u^l_i, v^l_j) \sim p(D_i, D_j)$ and negative if it is sampled from a product of marginals $(u^l_i, v^k_j) \sim p(D_i)p(D_j)$.
A single entity could be represented differently in dataset $D_i$ than in dataset $D_j$. More specifically, the digit "1" can be represented by an image in multiple domains, for example, as a handwritten digit in MNIST and a house number in the SVHN dataset. The same number $(1_{\text{MNIST}}, 1_{\text{SVHN})}$ chosen from MNIST and SVHN will be a positive pair, whereas a digit in MNIST paired with a different digit in SVHN is a negative pair (such as $(1_{\text{MNIST}}, 2_{\text{SVHN}})$).

The idea behind a critic function $f(u,v)$ is to assign higher values to positive pairs and lower values to negative pairs. The critic used in this study is a separable critic $f(u,v) = \frac{u^{\intercal} v}{\sqrt{d}}$, which is also used in the AMDIM~\cite{bachman2019amdim} implementation (e.g. there are other possible choices such as bilinear, concatenated critics~\cite{tschannen2019mutual}). Such a critic is equivalent to the scaled-dot product used in transformers~\cite{vaswani2017attention}.

Additionally, we clip scores from critic function $f(x,y)$ to interval $[-c, c]$ by $c\tanh(\frac{f(x,y)}{c})$ with $c=20$, like in AMDIM~\cite{bachman2019amdim}. Thus we need an additional term for a positive pair $e^{f^{c}(u^l_i,v^l_j)} = e^{-c}$. Lastly, we penalize $I(D_i; D_j)$ with the squared matching scores as $\lambda f(x,y)^2$ with $\lambda = 4\mathrm{e}{-2}$.

\subsection{Objectives}

Most contrastive self-supervised methods incorporate an estimator of mutual information. There are multiple ways of doing this, which are schematically shown in Figure~\ref{fig:dataset_scheme}. The combinations of objectives shown in Figure~\ref{fig:dataset_scheme} can be formulated as objectives based on Local DIM. In our taxonomy, we refer to the original Local DIM as CR since it captures a unimodal \emph{Convolution-to-Representation} relation. This method maximizes the mutual information between the location in the convolutional feature map $c_i$ (where the representation is considered to be along the channels) and the latent representation $z_i$. Further, AMDIM~\cite{bachman2019amdim}, ST-DIM~\cite{anand2019unsupervised}, and CM-DIM~\cite{sylvain2019locality} are referred to as XX, which captures a  \emph{Cross-Convolution-to-Representation} relation and as CC, which captures a \emph{Convolution-to-Convolution} relation. The XX objective pairs the latent $z_i$ with $c_j$, where $i \ne j$. The CC objective pairs $c_i$ with $c_j$, where $i \ne j$. The last type of pairing, RR, represents a \emph{Representation-to-Representation} relation and was introduced with CMC~\cite{tian2019contrastive} and SimCLR~\cite{chen2020simple}. These two methods pair $z_i$ and $z_j$, where $i \ne j$. The CR and XX objectives are more closely related to the Deep InfoMax principle, while the pairing of RR and CC is more closely related to CCA.

Researchers can combine these four basic contrastive objectives as shown in Figure~\ref{fig:variants} to create new objectives. Each combination of edges is a type of objective. The objective is equal to the total sum of all the edges. To create a full picture, we extend these objectives to other approaches such as autoencoders and DCCA~\cite{andrew2013deep} by adding the contrastive terms to their respective objective functions. We call these models the \emph{RR autoencoder} (RR-AE) and the \emph{CR canonical correlation analysis} (CR-CCA). We show them schematically in Figure~\ref{fig:others} alongside other baselines: a Supervised uni-modal model, a normal AE, a multimodal DCCAE~\cite{wang2015deep}, and a MMVAE~\cite{shi2019variational} with a loose \emph{ importance weighted autoencoder} (IWAE) estimator. We introduced the supervised model baseline specifically to serve as a discriminative bound on the multimodal dataset.

\section{Experiments}

\subsection{Datasets}
For our experiments, we incorporate diverse datasets with two input sources (shown in Figure~\ref{fig:dataset_scheme}). The first experiments are the most straightforward multi-view dataset, Two-View MNIST, and the more challenging multi-domain MNIST-SVHN dataset. These datasets allow us to validate our approach because the datasets gradually increase in multimodal dataset complexity. These datasets have the additional advantage of being tasks DCCAE, and MMVAE were directly evaluated in the original articles, allowing for a direct comparison to the author's reported results. The last dataset we evaluate is an Alzheimer's disease dataset, where we use a functional and structural view of the brain to learn representations that allow us to differentiate between healthy controls and patients with Alzheimer's disease.

\subsubsection{Multi-view dataset}
Two-View MNIST is inspired by~\cite{wang2015deep}, where each view represents a corrupted version of an original MNIST digit. First, the intensities of the original images are rescaled to the unit interval. The images are then resized to $32 \times 32$ to fit the DCGAN architecture. Lastly, to generate the first view, we rotate the image by a random angle within the $[-\pi/4, \pi/4]$ interval. For the second view, we add unit uniform noise and rescale the intensity to a unit interval again.

\subsubsection{Multi-domain dataset}
% TODO:clarify the above
The multi-domain dataset MNIST-SVHN was used by the authors of the MMVAE~\cite{shi2019variational}, where the first view is a grayscale MNIST digit and the second view is an RGB street view house number sampled from the SVHN dataset. The MNIST digits are modified by resizing the images to $32 \times 32$, which is also the default SVHN image size, to use them with the DCGAN encoder. All intensities are scaled to a unit interval. This dataset is more complicated than two-View MNIST because the digit is represented in different underlying domains. It is also more similar to the neuroimaging dataset because the views occur more naturally than the Two-View MNIST, where the views are augmentations of the original MNIST dataset.

\subsubsection{Multi-modal dataset}

The multimodal MRI dataset that we use is OASIS3~\cite{LaMontagne2019.12.13.19014902}. We use it to evaluate different representations for Alzheimer's disease (AD) classification. We use T1-weighted images to account for the anatomy of the brain. The T1-weighted image is brain masked using FSL~\cite{fsl} (v 6.0.2). T1 is the first modality and captures the structural aspects of the brain. The second modality captures the functional aspects of the brain. The second modality we use is resting-state fMRI (rs-fMRI), which captures the brain's metabolic function. We preprocess rs-fMRI into fALFF (relative low-frequency power in the 0.01 to 0.1 Hz power band) using REST~\cite{rest}.
All images are linearly converted to MNI space and resampled to 3mm isotropic voxel resolution. The final input volume is  $64\times64\times64$.
Careful selection (removing poor images and trying to limit race as a confounder) resulted in a final subset of the OASIS-3 dataset with 826 non-Hispanic Caucasian subjects. For each subject, we combined their sMRI and fALFF scans to create 4021 multimodal pairs. We left 100 ($66/22/12$) subjects for hold-out and used others in a stratified (about $70/15/15\%$) 5-folds for training and validation. We defined three groups: healthy cohort (HC), AD, and others (subjects with other brain problems). During pretraining, we employ all groups and pairs, whereas, during the linear evaluation, we only take one pair for each subject only use the HC or AD subjects. An additional preprocessing we applied is histogram standardization and z-normalization. During pretraining, we also use simple data augmentations, such as random crops and flips. These data augmentations were done using the TorchIO library~\cite{perez_garcia_torchio_2020}. The classes in the dataset are highly unbalanced, so we utilize a class balanced data sampler~\cite{balanced_data_sampler}.

\begin{figure}[t]
  \centering
  \includegraphics[width=\linewidth]{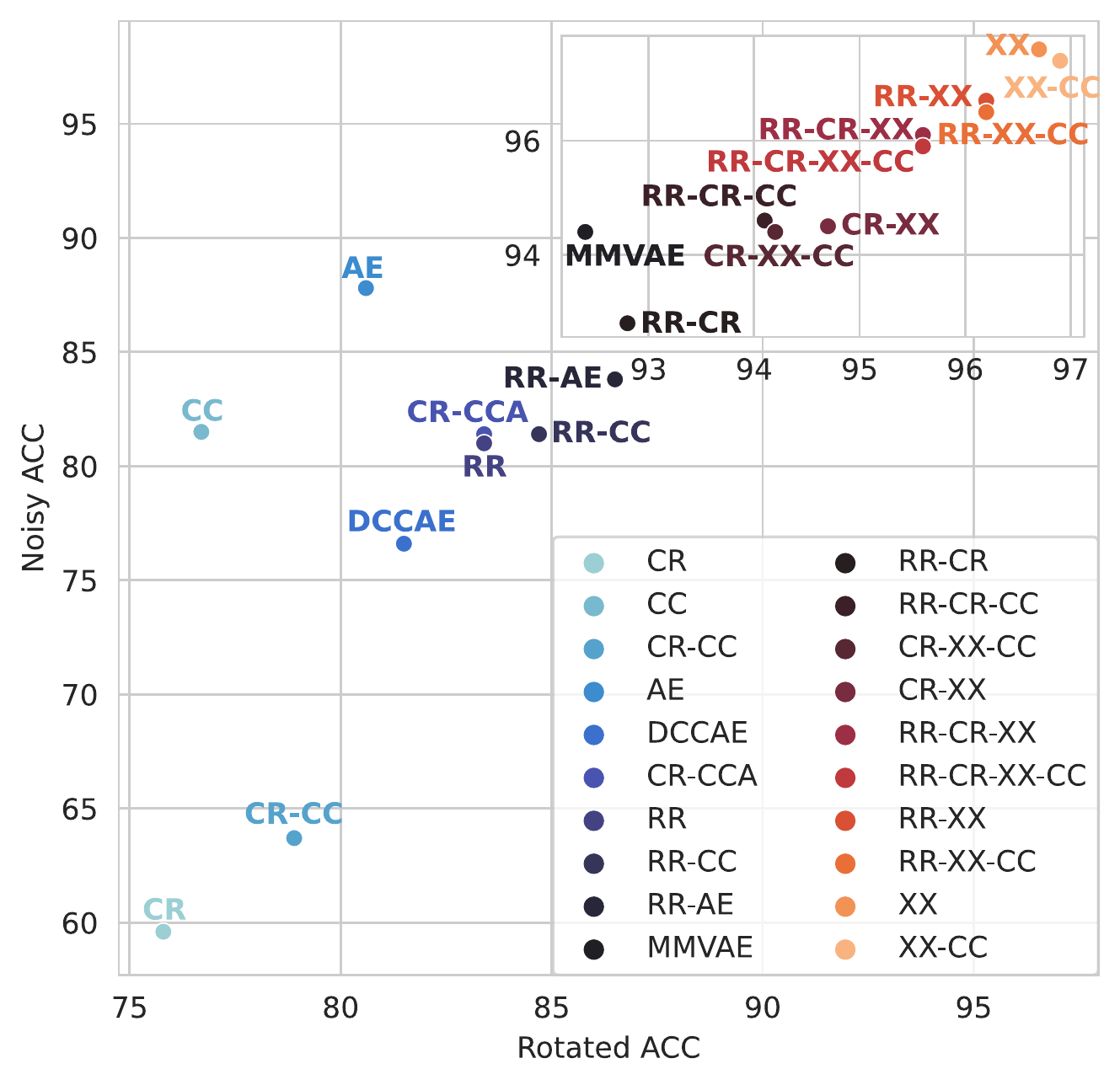}
  \caption{Test downstream perfomance with linear evaluation on Two-View MNIST. Cross-modal losses have a strong positive impact on performance.}
  \label{fig:two_view_mnist}
\end{figure}

\subsection{Evaluation}
\subsubsection{Linear evaluation on downstream task}
To evaluate representations on natural images, we employ the linear evaluation protocol, which is common for self-supervised approaches~\cite{bachman2019amdim,chen2020simple}. It trains a linear mapping from the latent representations to the number of classes. The weights in the encoder that produces these representations are kept frozen. In this study, we evaluate the encoder for each modality separately.

\subsubsection{Measuring similarity between representations}
To better understand the underlying inductive bias of the specific objectives, we measure the similarity between the representations of the different modalities it produces. The measure of similarity we use is \emph{canonical correlation analysis} CCA. CCA measures the average correlation of the aligned directions between the representations.  Additionally, we propose to use linear \emph{centered kernel alignment} (CKA)~\cite{kornblith2019similarity}, which has been shown to identify the relationship between representations of networks reliably.

Specifically, in our case, the CCA measure for a pair modalities $i$ and $j$ can be written as :
\begin{equation}
    CCA(Z^i, Z^j) = \frac{1}{d} ||Q^{\mathrm{T}}_{Z^j}Q_{Z^i}||_*,
\end{equation}
where $d$ is a dimension of latent representation, $Z$ is a $n \times d$ matrix of $d$-dimensional representation for $n$ samples, $||\cdot||_*$ is the nuclear norm, and $Q_{Z^i}$ is an orthonormal basis for $Z^i$.

The linear CKA measure is defined as:
\begin{equation}
CKA(Z^i, Z^j) = \left\|Z^{j\mathrm{T}} Z^i\right\|_{\mathrm{F}}^{2} /\left(\left\|Z^{i\mathrm{T}} Z^i\right\|_{\mathrm{F}}\left\|Z^{j\mathrm{T}} Z^j\right\|_{\mathrm{F}}\right),
\end{equation}
where $||\cdot||_F$ is the Frobenius norm.

\begin{figure}[t]
  \centering
  \includegraphics[width=\linewidth]{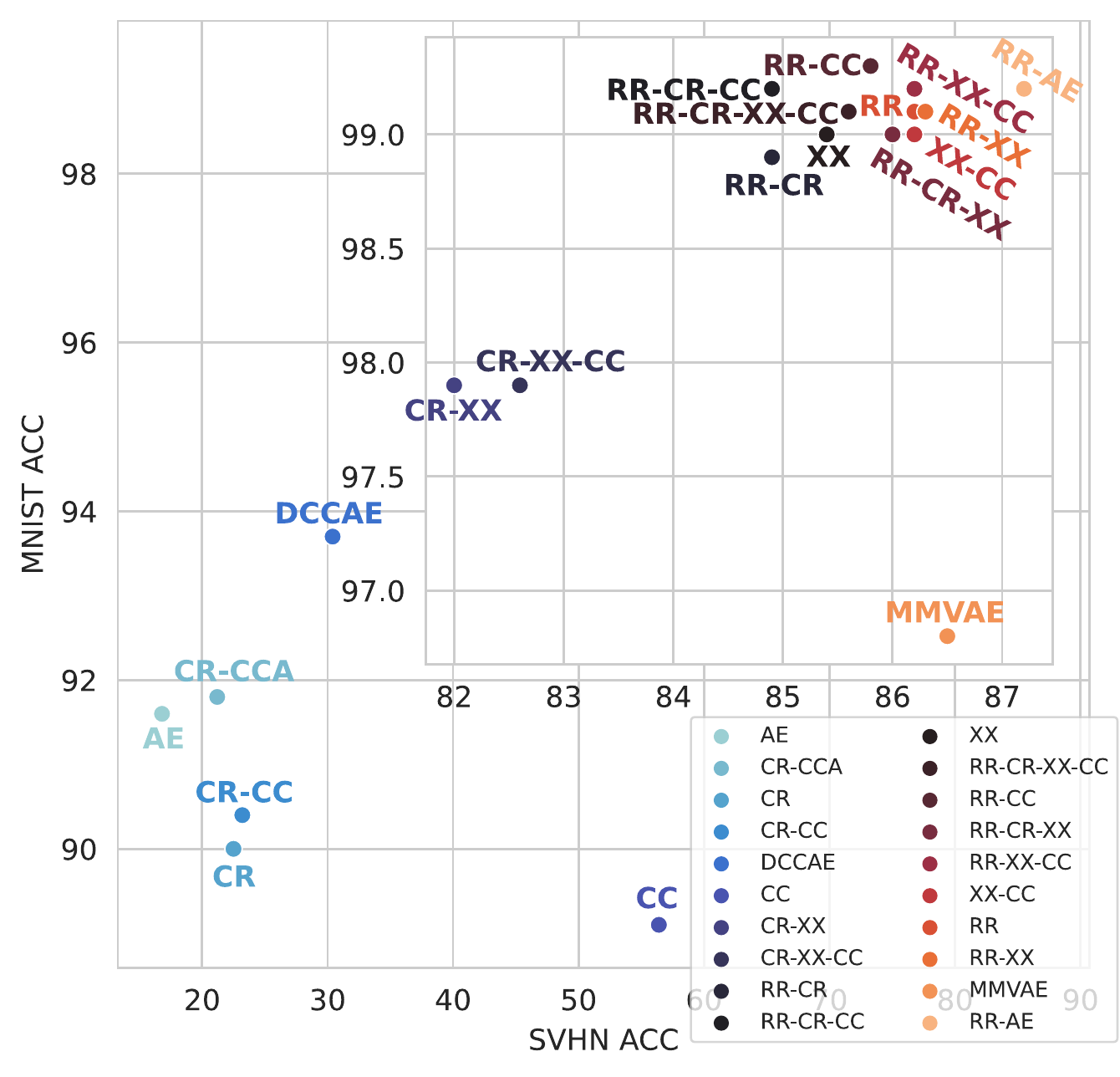}
  \caption{Downstream test performance with linear evaluation on MNIST-SVHN. We can see that overall, multimodal contrastive losses fare better than unimodal contrastive losses. }
  \label{fig:mnist_svhn}
\end{figure}

Additionally, we considered \emph{singular value CCA} (SVCCA)~\cite{NIPS2017_7188} and \emph{projected weighted CCA} (PWCCA)~\cite{morcos2018insights}. SVCCA is equivalent to CCA but performs an additional SVD-based dimensionality reduction. PWCCA is a weighted sum of the CCA vectors, where the weights are found through projection weighting. However, their results are similar to CCA.

\begin{figure}[t]
  \centering
  \includegraphics[width=\linewidth]{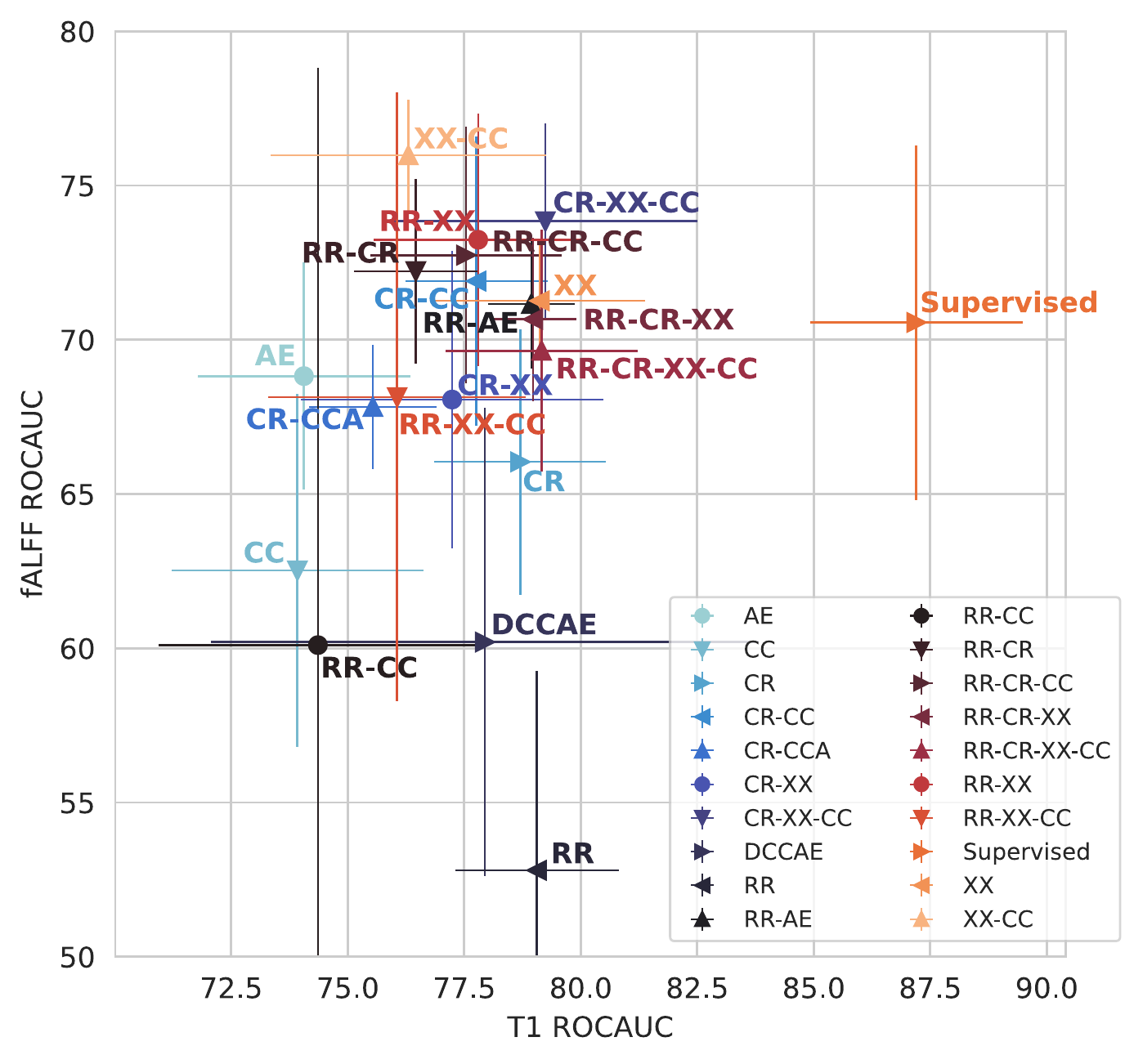}
  \caption{Holdout test downstream performance with linear evaluation on OASIS-3. The error lines show  $\pm$ standard deviation over models trained on 5 different folds.}
  \label{fig:oasis}
\end{figure}

\begin{figure*}[ht!]
  \centering
  \includegraphics[width=\linewidth]{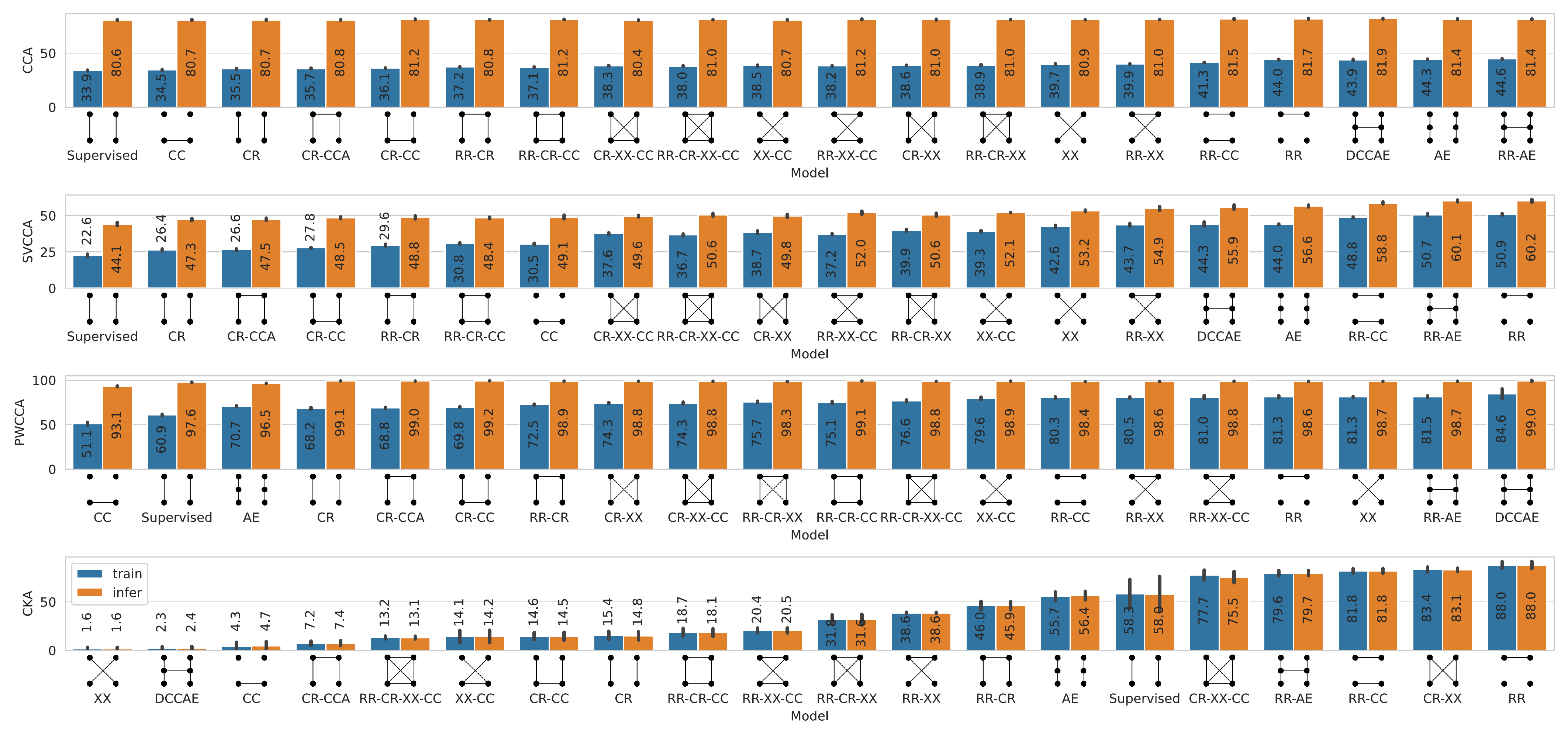}
  \caption{CCA, SVCCA, PWCCA and CKA measures on OASIS-3 for the training and holdout test sets.}
  \label{fig:oasis_svcca_cka}
\end{figure*}

\subsection{Implementation details}

The architecture and the hyper-parameters for each encoder and decoder are entirely based on DCGAN~\cite{radford2015unsupervised}. However, we removed one layer for experiments with natural images to be able to use an input size of $32x32$. The encoder produces a latent vector $z$, which in our case is $64$-dimensional. All the layers in the models are initialized with a uniform Xavier.

The projection head for the latent representation $z$ is identity. The convolutional features are taken from the layer with a feature side size of $8$ and passed through the convolutional projection head. The convolutional projection head is a ResNet like a block where one direction consists of $2$ convolutional layers (kernel size $1$, number of output and hidden features $64$, Xavier uniform initialization), and the second direction consists of one convolutional layer which (kernel size $1$, number of output features $64$, with initialization as identity). After summation of features from two directions, we apply a batch normalization layer to get the convolutional embeddings. The weights of the convolutional projection heads are shared across all contrastive objectives but separate for each modality.

The optimizer we use is RAdam~\cite{liu2019variance}($ \text{lr=}4\mathrm{e}{-4}$) with a OneCycleLR scheduler~\cite{smith2019super}($ \text{max\_lr=}0.01$). We pretrain on the natural images for $50$ epochs and on the volumes in OASIS-3 for $200$ epochs. The linear classification projection for natural images and OASIS-3 is trained for $50$ and $500$ epochs, respectively. All experiments were performed with a batch size of $64$. In some runs, we noticed that the CCA objective is unstable. The MMVAE did not converge for three folds out of 5 or completely collapsed due to the model's greater capacity. The greater capacity leads to GPU memory issues. Thus we had to decrease the batch size to 4. Thus we currently removed MMVAE from the OASIS-3 result.

The code was developed using PyTorch~\cite{NEURIPS2019_9015} and the Catalyst framework~\cite{catalyst}.
For data transforms of the brain images we utilized TorchIO~\cite{perez_garcia_torchio_2020}, for CKA analysis of the representations we used code by anatome~\cite{hataya2020anatome}, SVCCA~\cite{NIPS2017_7188}, for AMDIM~\cite{bachman2019amdim}, for DCCAE~\cite{perry2020mvlearn}, for MMVAE and MNIST-SVHN~\cite{shi2019variational}. The experiments were performed on an NVIDIA DGX-1 V100.

\section{Results}

\subsection{Two-View MNIST and MNIST-SVHN}
We can see in Figures~\ref{fig:two_view_mnist} and~\ref{fig:mnist_svhn} that cross-modal contrastive losses have a strong positive impact on downstream test performance across different architectural and model choices. We also note that the formulations have different performances across settings and datasets, leading to the conclusion that applying them in practice requires careful adaptation to a given problem.

Although contrastive methods will result in the best performance for a simple multi-view case, reconstruction-based models, such as MMVAE and RR-AA, stand out in multi-domain experiments. The performance for most of the contrastive multimodal models is within $3\%$ of RR-AE's performance. Thus one can choose decoder-free self-supervised approaches to reduce computational cost. It should also be noted that Uni-source AE, CR, multimodal DCCAE, and CR-CCA are not able to perform well on SVHN.

\subsection{OASIS-3 }

Figure~\ref{fig:oasis} shows the results on OASIS-3. Multimodal approaches exhibit strong performance, but the performance gain is less noticeable than with the previous tasks on natural images. Overall, accounting for robust performance on two modalities, the absolute leader in terms of the self-supervised methods is CR-XX-CC. The supervised baseline trained only on T1 modalities does still outperform this self-supervised method, however. In the case of the fALFF modality, the model that performs the best is XX-CC; notably, it is better than the Supervised model. This suggests that multimodal objectives can improve training for certain modalities that may be hard to extract discriminative representations from. Utilizing such an objective as some form of regularization may be beneficial.

We also want to note that the maximization of similarity between modalities by itself is not enough. The performance of the RR method shows that it might lead to a collapse of the representations. It might indicate that the T1 representation dominated over the fALFF representation during training because the RR objective model can learn meaningful T1 representations. By adding the reconstruction loss to the contrastive objective, we get RR-AE. The reconstruction loss improves the model. Analogous to the multi-domain experiment, the autoencoder is vital to learning the modality. Adding a reconstruction loss is not only the choice. The objective CR-XX-CC can, for example, be used, and it comes with reduced computational requirements because it does not require a decoder. The RR-AE model is also highly related to the idea and structure of the DCCAE. However, the typical implementation of CCA objectives in DCCAE is less numerically stable and requires the computation of eigenvalues and eigenvectors. Additionally, RR-AE is more robust and has a higher downstream performance.

The downstream performance might not be the main criteria to select a model, however. The multimodal models can be used to analyze the connection between modalities in neuroimaging~\cite{calhoun2016multimodal}. Thus one will want to have a model with a representation that has a shared subspace. Based on our experiments on downstream performance in Figure~\ref{fig:oasis} and similarity measures Figure~\ref{fig:oasis_svcca_cka} we would advise CR-XX-CC and RR-AE as models that perform well and should be investigated further.

\subsection{Understanding similarity measures}
Most self-supervised methods and the supervised model in Figure~\ref{fig:oasis_svcca_cka} are noticeably worse than DCCAE, RR, RR-AE, and AE in terms of the CCA and SVCCA metrics. The lower performance is, however, not found to be significant for the PWCCA metric.

Visually CCA, SVCCA, PWCCA measures of similarity behave comparably with a noticeable difference between training and testing sets. Interestingly, the CKA measure has very close values between training and testing sets. We hypothesize that the CKA measure shows the inductive bias hidden in representations through optimization, architecture, and the learned weights. The authors of the original CKA manuscript~\cite{kornblith2019similarity} argue that methods with higher CKA have a higher similarity between representations and that more of the subspace is shared. The reason why CCA, SVCCA, PWCCA behave poorly might be related to their sensitivity to noise.

\section{Conclusion}
In this work, we proposed a unifying view on contrastive methods and conducted a detailed study of their performance on multimodal datasets. We
believe that this unifying view will contribute to understanding how to learn powerful representations from multiple
modalities. Hopefully, instead of combining the similarities in various
ways and publishing the winning combinations as individual methods, the
the field will consider a broader perspective on the problem.

We empirically demonstrate that multimodal contrastive approaches result in performance improvements over methods that rely on a single modality for contrastive learning. We also show that downstream performance is highly dependent on the composition of such objectives. We argue that maximizing information similarity might not guarantee higher downstream performance. In some cases, it may weaken the representation or have a regularizing effect on the objective. However, high similarities between representations can be significant for other applications, i.e., multimodal analysis~\cite{calhoun2016multimodal}.

DIM-based methods have a smaller computational cost than autoencoder-based methods because they do not require a decoder to be trained. The smaller computational cost lowers the hardware requirements. While DIM-based methods do have comparable downstream performance, the lower hardware requirements can help democratize medical imaging. DIM-based methods can also be helpful in cases where labels do not exist or are inaccurate, a scenario that is quite common in neurological and mental disorder nosology.
% TODO expand here.

For future work, we are interested in considering how the conclusions we draw here hold in different learning settings with scarcer data or annotations such as few-shot or zero-shot learning cases. Another goal is to study the joint shared subspace projects in brain space for visualization. The proposed interpretable joint learning approach can help advance work in our search for neuroimaging biomarkers.

\section{Acknowledgments}
\label{sec:acknowledgments}
This work is supported by NIH R01 EB006841.

Data were provided in part by OASIS-3: Principal Investigators: T. Benzinger, D. Marcus, J. Morris; NIH P50 AG00561, P30 NS09857781, P01 AG026276, P01 AG003991, R01 AG043434, UL1 TR000448, R01 EB009352. AV-45 doses were provided by Avid Radiopharmaceuticals, a wholly-owned subsidiary of Eli Lilly.

\bibliographystyle{IEEEbib}
\bibliography{references}

\end{document}